\documentclass[runningheads]{llncs}
\usepackage[T1]{fontenc}

\usepackage{float}
\usepackage{amssymb}
\usepackage{multirow}
\usepackage{booktabs}
\usepackage{graphicx}
\usepackage{subcaption}
\usepackage{amsmath}
\usepackage{placeins}
\usepackage{orcidlink}
\usepackage[ruled,vlined]{algorithm2e}
\begin{document}
\title{DualGate-Net: A Prior-Gated Dual-Encoder Framework for Histopathology Cell Detection}
\titlerunning{DualGate-Net for Cell Detection}
\author{
Bahman Jafari Tabaghsar\inst{1}\orcidlink{0009-0008-8592-9391} \and
Son Tran\inst{1}\orcidlink{0000-0002-5912-293X}
\and K. Devaraja\inst{2}\and
Atul Sajjanhar\inst{1}\orcidlink{0000-0002-0445-0573}
}

\authorrunning{Jafari Tabaghsar et al.}

\institute{
School of Information Technology, Deakin University, Burwood, VIC 3125, Australia\\
\email{\{b.jafaritabaghsar,atul.sajjanhar, son.tran\}@deakin.edu.au}
\and
Kasturba Medical College, Manipal Academy of Higher Education, Manipal, Karnataka 576104, India\\
\email{devaraja.k@manipal.edu}
}
\maketitle              
\begin{abstract}
Cell detection in histopathology images strongly depends on surrounding tissue context, where visually similar cells may belong to different classes under different microenvironments. Recent tissue-aware methods incorporate contextual priors, but often rely on static fusion strategies that may propagate noisy information. In this work, we propose DualGate-Net, a prior-aware dual-encoder framework that combines a ConvNeXtV2-based local encoder and a SegFormer-based global encoder through a learnable prior-gated fusion mechanism. The proposed module adaptively regulates the influence of tissue priors across spatial locations, while an auxiliary foreground reconstruction branch preserves high-frequency cellular structures during training. In addition, auxiliary cellness-guided cues are incorporated to further improve localization robustness. Experiments on the OCELOT benchmark demonstrate consistent improvements, achieving macro F1-scores of 0.7722 on the validation set and 0.7345 on the test set, highlighting the effectiveness of adaptive prior integration for robust histopathology cell detection.
\end{abstract}

\section{Introduction}
Computational pathology (CPATH) extends digital pathology by employing artificial intelligence techniques for the analysis of digitized tissue specimens, particularly whole-slide images (WSIs), in clinical workflows~\cite{CPATH}. Despite significant progress, accurate cell (or nucleus) detection plays a fundamental role in enabling a wide range of applications, including cancer diagnosis, grading, prognosis estimation, and treatment planning~\cite{graham2019hover,wu2023multi,ryu2023ocelot}. In clinical practice, pathologists address this challenge by analyzing WSIs at multiple magnification levels, first capturing global tissue architecture and then focusing on fine-grained cellular details, highlighting the importance of jointly modeling local and contextual information for robust cell detection~\cite{ryu2023ocelot,shui2026towards}.
Recent advances in improved automated cell detection, tissue-aware approaches (OCELOT)~\cite{ryu2023ocelot} have demonstrated that incorporating contextual information can enhance detection performance by providing additional cues about the spatial distribution of cancerous regions. Despite these advances, several important challenges remain unresolved. Unlike conventional segmentation-based datasets~\cite{Puma,pannuk,MoNuSeg}, OCELOT formulates cell detection as a centroid-based classification problem, making accurate localization substantially more difficult in dense and overlapping cellular regions. Furthermore, tissue priors used in context-aware frameworks are typically generated from auxiliary prediction models and therefore inherently contain uncertainty and noise. However, existing methods often incorporate such priors through static fusion or direct concatenation strategies ~\cite{ryu2023ocelot,horst2024cellvit}, which can propagate unreliable contextual signals and negatively affect feature learning. In addition, current CNN-based approaches struggle to capture long-range contextual dependencies~\cite{ryu2023ocelot,horst2024cellvit}, while transformer-based methods~\cite{miler,horst2024cellvit}, despite their strong global modeling capability, often lack an adaptive mechanism for jointly leveraging fine-grained local morphology and noisy tissue-level contextual priors.
To address these limitations, we propose a prior-aware dual-encoder framework for cell detection that explicitly models both local morphological features and global contextual information. The proposed architecture combines a ConvNeXtV2-based~\cite{convenext} local encoder with a SegFormer-based~\cite{segformer} global encoder, enabling complementary feature extraction across multiple scales. To effectively incorporate prior information, we introduce a learnable prior-gated fusion mechanism that adaptively controls the influence of tissue-derived priors at each spatial location, allowing the model to selectively refine and utilize contextual signals. In addition, we incorporate an auxiliary foreground reconstruction objective to encourage the preservation of high-frequency cellular structures, and further enhance localization performance by integrating a class-agnostic cellness prior derived from a teacher model.

Experimental results on the OCELOT benchmark show that the proposed framework achieves robust and consistent improvements in cell detection performance across challenging pathological regions. The main contributions of this work can be summarized as follows: (1) we propose a dual-encoder architecture that effectively integrates local and global representations for cell detection; (2) we introduce a prior-gated fusion mechanism for adaptive and spatially-aware integration of tissue priors; (3) we incorporate auxiliary supervision and cellness-guided signals to improve robustness and localization accuracy; and (4) extensive experiments on OCELOT and additional cross-dataset analysis on BRCA demonstrate the effectiveness and generalization capability of the proposed adaptive prior integration framework.
\section{Related Work}
\subsection{CNN-based Cell Detection Methods}
Initial studies on histopathology cell detection were largely based on convolutional neural network (CNN) architectures and semantic segmentation frameworks for local morphology modeling. These approaches generally formulate cell detection as dense probability-map prediction or pixel-wise segmentation followed by centroid extraction through post-processing operations such as non-maximum suppression or watershed algorithms~\cite{graham2019hover}. Within the OCELOT challenge, the official baseline adopted a DeepLabV3+-based framework trained on high-resolution cell patches using Gaussian-based point supervision~\cite{ryu2023ocelot}. Several competition submissions further extended CNN-based tissue-aware learning paradigms. For example, SoftCTM incorporated soft Gaussian label generation and tissue-aware contextual learning into a DeepLabV3+ framework~\cite{softctm}, while FC-HarDNet-based approaches utilized auxiliary tissue segmentation priors to refine cell classification predictions~\cite{hardnet}. Other methods employed ensemble CNN architectures based on ResNet-50 for robust local feature extraction~\cite{resnet50}. 
Although CNN-based methods demonstrate strong capability in capturing fine-grained nuclear morphology and texture information, their limited receptive fields restrict effective modeling of large-scale tissue context and long-range spatial dependencies. Consequently, these approaches often struggle in challenging pathological regions where local cellular appearance alone is insufficient for reliable classification.
\subsection{Transformer-based Cell Detection Methods}

To address the limitations of purely local modeling, recent studies have increasingly explored transformer-based and global context-aware architectures for histopathology analysis. Inspired by the multi-scale diagnostic workflow of pathologists, these methods aim to jointly model fine-grained cellular structures together with broader tissue organization~\cite{shui2026towards}. The OCELOT benchmark demonstrated that incorporating tissue-level contextual priors can significantly improve cell detection performance by providing complementary information about the spatial distribution of cancerous regions~\cite{ryu2023ocelot}. 
Building upon this observation, several OCELOT challenge submissions adopted transformer-based architectures to improve contextual reasoning in whole-slide images. CellViT employed vision transformer representations for joint cell segmentation and classification, demonstrating improved contextual feature learning compared to conventional CNN-based frameworks~\cite{horst2024cellvit}. Similarly, SegFormer-based architectures showed strong capability in capturing long-range dependencies while maintaining efficient hierarchical feature extraction~\cite{segformer}. S{\ae}ther \textit{et al.} further proposed a transformer-based additive joint pred-to-decoder framework that injects tissue predictions into intermediate decoder stages for improved tissue-aware cell detection on the OCELOT benchmark~\cite{ha2023generating}. In parallel, recent hybrid context-aware frameworks combining convolutional and transformer-based representations demonstrated the complementary benefits of local morphological modeling and global contextual reasoning in histopathology analysis~\cite{segnet}.
Despite these advances, several limitations remain unresolved. Existing context-aware approaches often rely on static fusion or direct concatenation when integrating tissue priors, without considering the uncertainty and noise associated with predicted contextual information~\cite{ryu2023ocelot,horst2024cellvit}. As a result, unreliable tissue predictions may negatively affect feature learning and localization performance. Furthermore, although transformer-based architectures improve global context modeling, excessive reliance on high-level contextual representations may suppress fine-grained cellular morphology that remains critical for accurate cell classification in dense and overlapping regions.
Motivated by these limitations, our work introduces a prior-aware dual-encoder framework that adaptively integrates CNN-based local morphological representations and transformer-based global contextual features through learnable gated fusion mechanisms. Furthermore, we incorporate auxiliary foreground supervision and cellness-guided priors to improve robustness, preserve fine-grained cellular structures, and enhance localization performance in challenging and densely populated regions.

\section{Method}
\subsection{Baseline Architecture}
We adopt a dual-encoder architecture with learnable fusion to jointly capture complementary representations from histopathology images. Specifically, a local encoder is used to extract fine-grained morphological features, while a global encoder models long-range contextual dependencies.

Such dual-branch designs have been shown to be effective for combining local and global information in medical image analysis~\cite{segnet}. However, in our preliminary experiments, we observe that simply combining a convolutional encoder (ResNet34) with a transformer-based global encoder (SegFormer) provides only marginal performance improvements. This observation suggests that the effectiveness of the dual-encoder framework is highly dependent on the representation capacity of the local encoder, particularly for capturing subtle cellular structures and improving generalization to unseen data.

Motivated by this, we replace the conventional convolutional backbone with a more advanced ConvNeXtV2-based encoder, which provides stronger feature representations and improved robustness. As demonstrated in our experiments, this modification leads to significant performance gains, especially on validation and test sets, indicating improved generalization capability.
Formally, let $U_l$ and $S_l$ denote the local and global feature maps at scale $l$, respectively. These features are spatially aligned and fused before being passed to the decoder.
\begin{figure}[!t]
    \centering
    \includegraphics[width=0.8\textwidth]{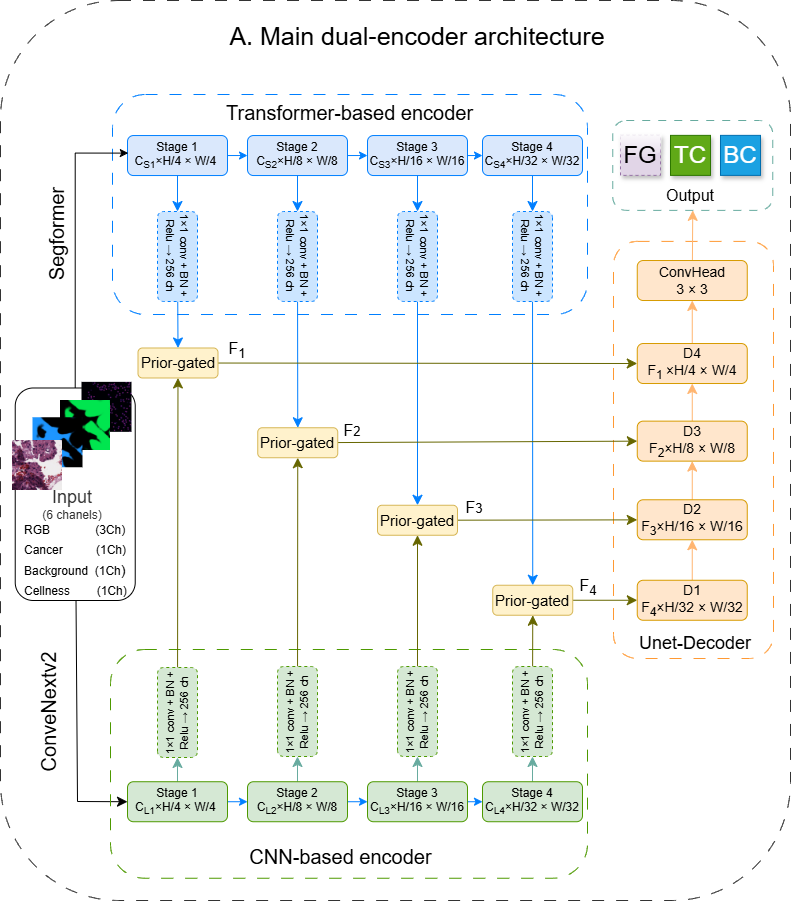}
    \caption{Overview of the proposed DualGate-Net architecture. The framework combines a ConvNeXtV2-based local encoder and a SegFormer-based global encoder through multi-scale prior-gated fusion modules. The decoder jointly predicts tumor cells (TC), background cells (BC), and auxiliary foreground reconstruction (FG). Detailed structures of the prior-gated fusion module and auxiliary FG (foreground reconstruction) branch are illustrated in Fig.~\ref{fig:prior_gate} and Fig.~\ref{fig:auxiliary_branch}, respectively.}
    \label{fig:main_architecture}
\end{figure}
\FloatBarrier
\subsection{Prior Generation}
Given the strong dependency between tissue context and cell-level predictions in histopathology images~\cite{ryu2023ocelot}, we first train a SegFormer-B2 model for tissue segmentation to provide contextual information.

In ~\cite{miler}, they extract the softmax output corresponding to the cancer class to construct a single-channel cancer-area probability map, which serves as a coarse spatial prior. In our extended formulation, we further exploit the full soft tissue predictions by constructing multi-channel priors, where each channel corresponds to a specific tissue class (cancer and background). These class-wise probability maps are concatenated with the input and provide richer contextual guidance to the proposed model.
\subsection{Prior-Gated Multi-Scale Fusion}
Existing OCELOT-based approaches typically incorporate tissue-derived priors using static fusion strategies, such as direct concatenation or additive injection~\cite{horst2024cellvit,ryu2023ocelot}. However, predicted tissue priors may contain uncertainty and spatially varying noise. Treating all prior responses uniformly can therefore propagate unreliable contextual signals into the cell detection branch. To address this limitation, we propose a prior-gated fusion module that adaptively regulates the contribution of prior information at each spatial location and feature scale.

\begin{figure}[!t]
    \centering
    \includegraphics[width=0.80\textwidth]{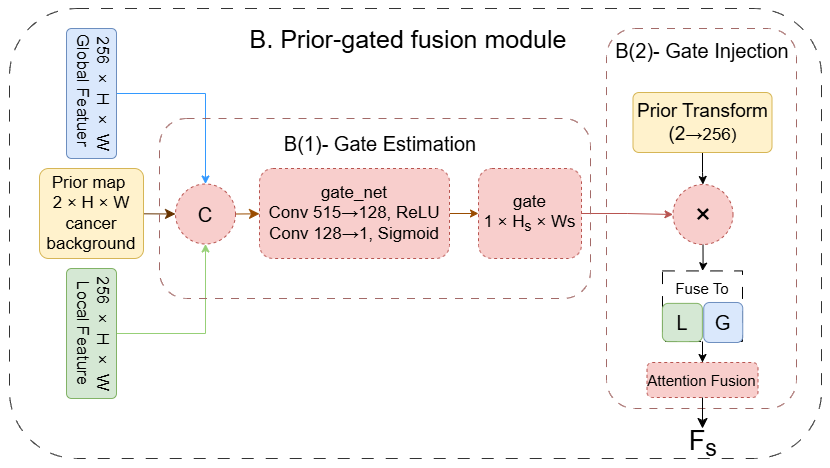}
    \caption{
    Prior-gated fusion module. B(1) estimates a spatial reliability gate from the concatenation of local features, global features, and tissue prior maps. B(2) projects the prior maps into the local and global feature spaces, modulates them using the learned gate, and performs attention-based fusion to obtain the final fused representation.
    }
    \label{fig:prior_gate}
\end{figure}

As illustrated in Fig.~\ref{fig:prior_gate}, the proposed module consists of two main stages. In the first stage, denoted as B(1), the module estimates a spatial reliability gate from the local features, global features, and prior maps. At each scale $l$, let $U_l$ and $S_l$ denote the local CNN-based and global transformer-based feature maps, respectively, and let $P_l$ denote the resized prior map at the same spatial resolution. The gate is computed as:
\begin{equation}
g_l = \sigma \left( W_{g2} * \phi \left( \mathrm{BN}(W_{g1} * [U_l; S_l; P_l]) \right) \right),
\end{equation}
where $[\cdot;\cdot]$ denotes channel-wise concatenation, $W_{g1}$ and $W_{g2}$ are convolutional layers, $\mathrm{BN}$ denotes batch normalization, $\phi$ is the GELU activation function, and $\sigma$ is the sigmoid function. The resulting gate $g_l$ assigns spatially adaptive weights to prior information, allowing the model to suppress unreliable prior responses while emphasizing informative contextual regions.

In the second stage, denoted as B(2), the prior map is projected into the feature spaces of both encoder streams and modulated by the learned gate:
\begin{equation}
\hat{U}_l = U_l + g_l \odot (W_u * P_l),
\end{equation}
\begin{equation}
\hat{S}_l = S_l + g_l \odot (W_s * P_l),
\end{equation}
where $W_u$ and $W_s$ are projection layers for the local and global branches, respectively, and $\odot$ denotes element-wise multiplication. The gated prior-enhanced features are then concatenated and refined using an attention-based fusion module:
\begin{equation}
F_l = \mathrm{CBAM}([\hat{U}_l; \hat{S}_l]).
\end{equation}
This design enables spatially adaptive prior integration across multiple scales. Instead of injecting tissue priors uniformly, the proposed module learns where and how strongly prior information should influence the local and global feature representations.

\begin{algorithm}[!t]
\caption{Prior-Gated Multi-Scale Fusion}
\KwIn{Local features $\{U_l\}_{l=1}^{L}$, global features $\{S_l\}_{l=1}^{L}$, prior map $P$}
\KwOut{Fused features $\{F_l\}_{l=1}^{L}$}

\For{$l = 1$ \KwTo $L$}{
    $P_l \leftarrow \mathrm{Resize}(P)$\;
    $X_l \leftarrow \mathrm{Concat}(U_l, S_l, P_l)$\;
    $g_l \leftarrow \sigma \left( W_{g2} * \phi \left( \mathrm{BN}(W_{g1} * X_l) \right) \right)$\;

    $P_l^{U} \leftarrow W_u * P_l$\;
    $P_l^{S} \leftarrow W_s * P_l$\;

    $\hat{U}_l \leftarrow U_l + g_l \odot P_l^{U}$\;
    $\hat{S}_l \leftarrow S_l + g_l \odot P_l^{S}$\;

    $F_l \leftarrow \mathrm{CBAM}(\mathrm{Concat}(\hat{U}_l, \hat{S}_l))$\;
}
\Return{$\{F_l\}_{l=1}^{L}$}
\end{algorithm}

\FloatBarrier
\subsection{Auxiliary Foreground Reconstruction}
\begin{figure}[ht]
    \centering
    \includegraphics[width=0.80\textwidth]{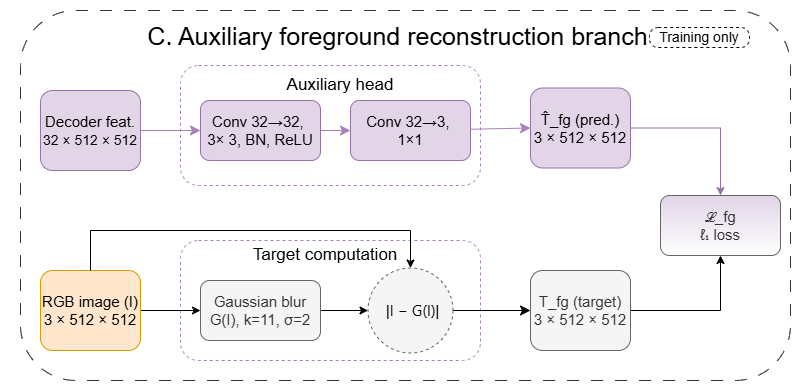}
    \caption{
   Architecture of the auxiliary foreground reconstruction branch used during training. The module reconstructs foreground-enhanced RGB representations from decoder features to preserve high-frequency cellular structures and improve localization robustness.
    }
    \label{fig:auxiliary_branch}
\end{figure}

To further preserve fine-grained cellular morphology and high-frequency structural information, we introduce an auxiliary foreground reconstruction branch that operates on the decoder features during training. As illustrated in Fig.~\ref{fig:auxiliary_branch}, the auxiliary branch encourages the network to retain discriminative structural details such as cell boundaries, edges, and foreground texture patterns that are often degraded during deep feature abstraction.
Instead of directly reconstructing the original RGB image, we define a foreground-enhanced reconstruction target using a high-frequency residual representation:
\begin{equation}
T_{\text{fg}} = \left| I - G(I) \right|,
\end{equation}
where $I$ denotes the input RGB image and $G(\cdot)$ represents Gaussian smoothing. The subtraction operation suppresses low-frequency background information while emphasizing structurally informative foreground regions and cellular boundaries.
The auxiliary branch receives decoder features and predicts a reconstructed foreground representation $\hat{T}_{\text{fg}}$ through a lightweight convolutional head. The reconstruction objective is defined using the $\ell_1$ loss:
\begin{equation}
\mathcal{L}_{\text{fg}} = \| \hat{T}_{\text{fg}} - T_{\text{fg}} \|_1.
\end{equation}
Our auxiliary foreground reconstruction branch is designed to preserve fine-grained cellular morphology while the model exploits tissue priors. This auxiliary supervision regularizes feature learning by encouraging the decoder to preserve high-frequency structural cues throughout training. Consequently, the network becomes more robust in dense and ambiguous pathological regions where accurate boundary preservation is critical for reliable cell localization. Importantly, the auxiliary branch is only used during training and is removed during inference, introducing no additional computational overhead at test time.

\FloatBarrier
\subsection{Training Objective}
The proposed model is trained using a regression-based objective, where the network predicts dense heatmaps for cell localization. Specifically, the main decoder outputs class-specific heatmaps corresponding to cancer and background cell regions.
Given the predicted heatmaps $Z$ and the ground-truth heatmaps $Y$, where the ground-truth maps are generated using Gaussian kernels centered at annotated cell locations. the primary loss is defined as a mean squared error (MSE) loss:
\begin{equation}
\mathcal{L}_{\text{seg}} = | Z - Y |_2^2
\end{equation}
In addition to the main objective, an auxiliary foreground reconstruction loss is applied at the bottleneck level, as defined in Eq.~(6). The overall training objective is formulated as:
\begin{equation}
\mathcal{L} =
\mathcal{L}_{\text{seg}}
+
\lambda \mathcal{L}_{\text{fg}}
\end{equation}

where $\lambda$ is a weighting factor that balances the contribution of the auxiliary loss.

This formulation encourages the model to jointly learn accurate cell localization through heatmap regression while preserving fine structural details via the auxiliary high-frequency reconstruction objective. The model is optimized using the Adam optimizer under standard training settings.

\section{Experiments and Results}
\subsection{Dataset}
\begin{figure}[ht]
\centering
\includegraphics[width=0.90\linewidth]{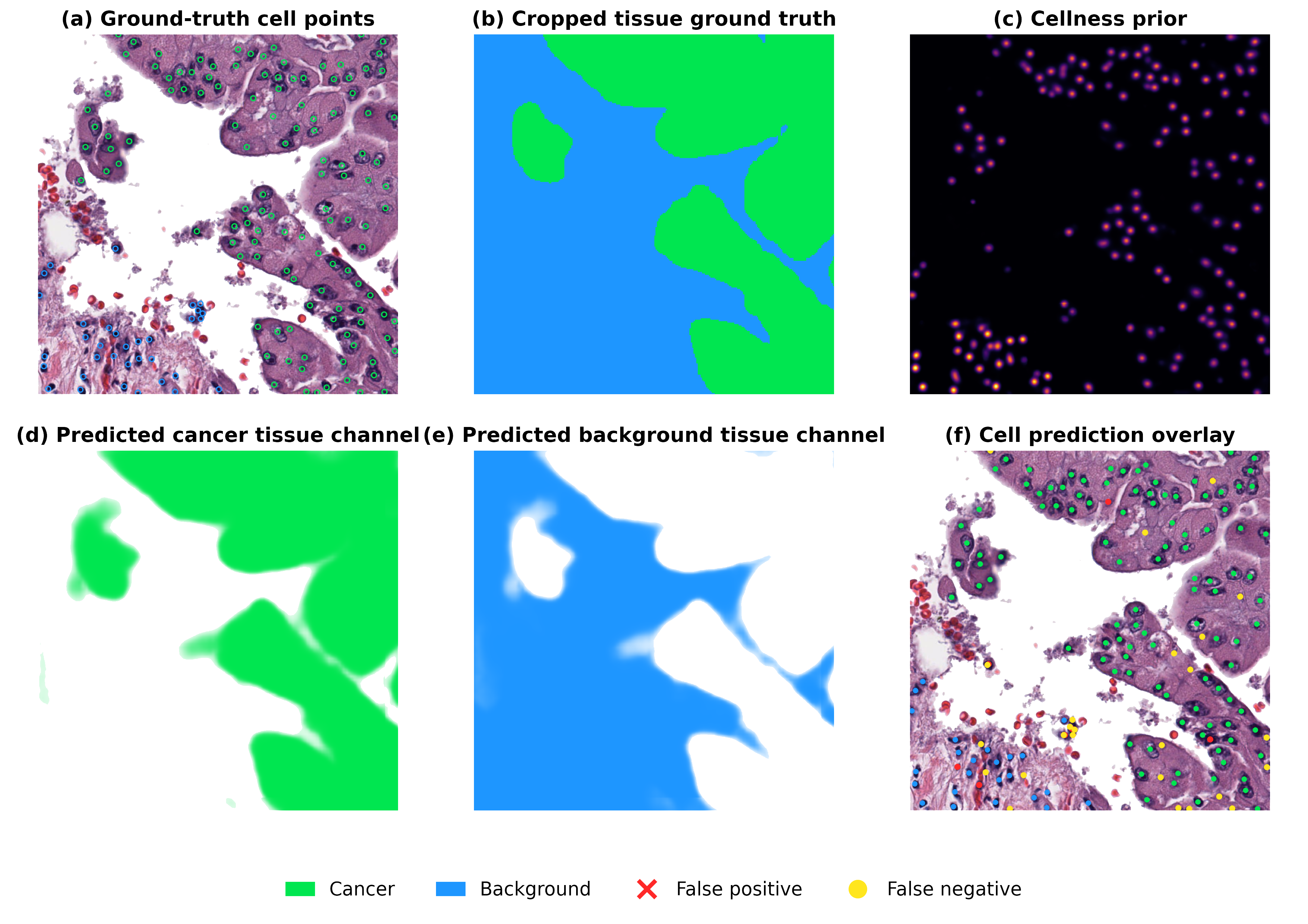}
\caption{
Example visualization from the OCELOT benchmark.
(a) Ground-truth cell centroid annotations.
(b) Tissue-level cancer/background annotations.
(c) Generated cellness prior.
(d,e) Predicted tissue prior channels.
(f) Final prediction overlay with false positives and false negatives.
}
\label{fig:dataset_example}
\end{figure}

We used the OCELOT 2023 dataset (version 1.0.1)~\cite{ryu2023ocelot}, a paired histopathology benchmark designed for tissue-aware cell analysis. The dataset contains 663 paired samples split into 400 training, 137 validation, and 126 test cases. Each sample includes a high-resolution cell patch and a corresponding larger-field-of-view tissue patch extracted from the same TCGA whole-slide image. Cell annotations are provided as centroid coordinates with binary labels for background cells (BC) and tumor cells (TC), while tissue annotations are provided as pixel-wise masks for cancer and background regions. Unlike conventional cell detection datasets, OCELOT explicitly models the interaction between tissue-level context and cell-level localization. Fig.~\ref{fig:dataset_example} illustrates an example OCELOT sample, including cell annotations, tissue supervision, generated cellness priors, predicted tissue channels, and final prediction overlays.

\subsection{Implementation Details}

The proposed framework is trained as a heatmap regression model using class-specific Gaussian targets. The primary objective is a mean squared error (MSE) loss between predicted and ground-truth heatmaps, while configurations with the auxiliary foreground reconstruction branch additionally employ an L1 reconstruction loss at the bottleneck level. 
Detailed experimental settings, including network configuration, optimization parameters, input channels, and augmentation strategies, are summarized in Table~\ref{tab:experimental_config}.

\begin{table}[!t]
\centering
\caption{Experimental configuration.}
\label{tab:experimental_config}
\footnotesize
\setlength{\tabcolsep}{5pt}
\begin{tabular}{ll}
\hline
\textbf{Configuration} & \textbf{Value} \\
\hline
Input size & 512 $\times$ 512 \\
Input channels & 6 (RGB + priors + cellness) \\
Global encoder & SegFormer-B2 \\
Local encoder & ConvNeXtV2-Tiny \\
Decoder & U-Net style (256, 128, 64, 32) \\
Epochs & 250 \\
Batch size & 4 \\
Optimizer & AdamW \\
Learning rate & $1 \times 10^{-5}$ \\
LR schedule & Polynomial decay \\
Main loss & MSE (heatmap regression) \\
Auxiliary loss & L1 (foreground reconstruction) \\
Auxiliary weight & 0.005 \\
Augmentation & Flip, crop, blur, color jitter \\
\hline
\end{tabular}
\end{table}
\FloatBarrier
\FloatBarrier
\subsection{Cellness Prior}
In addition to tissue-derived priors, we explore the use of a class-agnostic cellness prior as an auxiliary input signal to improve localization performance. Inspired by prior works that leverage cellness or objectness cues for enhancing detection robustness~\cite{miler}, we construct a cellness map using a teacher model by taking the maximum response over tumour-cell and background-cell heatmaps.

The resulting cellness prior is incorporated as an additional input channel, leading to a 6-channel input configuration when combined with RGB and tissue priors. This configuration is evaluated in our experiments (see Table~\ref{tab:ablation}), where the performance gains associated with the 6-channel setting reflect the contribution of the cellness prior.

Importantly, this prior is not fused multiplicatively with tissue priors, but rather used independently and jointly processed within the proposed framework to avoid disrupting the contextual information derived from tissue segmentation.

While not a primary contribution, the inclusion of cellness information serves as a practical enhancement, enabling the model to better localize cells in challenging regions with ambiguous tissue context.


\subsection{Main Results}
As shown in Table~\ref{tab:main_results}, the proposed method achieves the best overall performance on both the validation and test sets. On the validation set, our model obtains a macro F1-score of 77.22

In addition to the overall improvement, the proposed method consistently enhances performance across both tumor cell (TC) and background cell (BC) categories, demonstrating robust behavior under varying tissue contexts. The improvements observed on the validation set are effectively transferred to the test set, suggesting strong generalization capability without overfitting.

Compared to existing state-of-the-art approaches, the performance gains highlight the effectiveness of combining complementary feature representations with adaptive prior integration. These results indicate that leveraging both local morphological cues and global contextual information, together with structured prior guidance, plays a crucial role in achieving accurate and reliable cell detection in challenging histopathology images.

\subsection{Ablation Study}
We conduct a series of ablation experiments to evaluate the contribution of each component in the proposed framework, as summarized in Table~\ref{tab:ablation}. In this context, the \emph{Ch} column denotes the number of input channels corresponding to prior information, where tissue-derived priors (cancer and background) and the class-agnostic cellness prior are incorporated as separate channels. In particular, the 6-channel configuration corresponds to the inclusion of the additional cellness prior alongside the RGB image and tissue priors.

Starting from the baseline ACS-SegNet model, incorporating tissue-aware priors and the proposed prior-gated fusion mechanism consistently improves performance, highlighting the importance of adaptive contextual integration for robust cell detection. Introducing the auxiliary foreground reconstruction branch further improves both validation and test performance, demonstrating its effectiveness in preserving high-frequency structural details and enhancing localization robustness in dense cellular regions.

Increasing the number of input channels from 5 to 6 by incorporating the additional cellness prior leads to further gains, suggesting that the teacher-guided cellness signal provides complementary localization cues beyond tissue-level contextual information.
Furthermore, replacing the baseline architecture with the proposed dual-encoder framework consistently improves performance across all experimental settings, confirming the effectiveness of jointly modeling local morphological features and global contextual representations. Finally, integrating all proposed components yields the best overall performance, achieving macro F1-scores of 77.22 on the validation set and 73.45 on the test set.
\begin{table}[ht]
\centering
\caption{Comparison with prior methods on the OCELOT dataset.}
\label{tab:main_results}
\footnotesize
\setlength{\tabcolsep}{12pt}
\resizebox{\linewidth}{!}{
\begin{tabular}{|l|ccc|ccc|}
\hline
\textbf{Method} 
& \multicolumn{3}{c|}{\textbf{Val}} 
& \multicolumn{3}{c|}{\textbf{Test}} \\
& TC & BC & mF1 & TC & BC & mF1 \\
\hline
Li et al. & -- & -- & 75.18 & 77.53 & 67.35 & 72.44 \\
Millward et al. & 80.73 & 68.72 & 74.73 & 76.87 & 67.55 & 72.21 \\
SoftCTM & -- & -- & 70.46 & 76.11 & 67.34 & 71.89 \\
OCELOT & -- & -- & 72.68 & 74.64 & 67.73 & 71.23 \\
HardNet & 78.60 & 65.04 & 71.82 & 74.57 & 65.27 & 69.92 \\
U-Net & -- & -- & 65.60 & -- & -- & 62.38 \\
ACS-SegNet & 81.01 & 68.95 & 74.98 & 76.60 & 67.82 & 72.21 \\

\hline
\textbf{Ours} & \textbf{83.87} & \textbf{71.93} & \textbf{77.22} & \textbf{78.16} & \textbf{68.74} & \textbf{73.45} \\
\hline

\end{tabular}
}
\end{table}
\begin{table}[ht]
\centering
\caption{Ablation study of different components.}
\label{tab:ablation}
\footnotesize
\setlength{\tabcolsep}{6pt}
\begin{tabular}{|p{2.2cm}|c|c|c|c|c|c|}
\hline
\centering\textbf{Model} & \textbf{Ch} & \textbf{Gate} & \textbf{FG} & \textbf{Cellness} & \textbf{Val137} & \textbf{Test126} \\
\hline
\multirow{4}{*}{\centering ACS-SegNet} 
& 3 & -- & -- & -- & 70.93 & 68.88 \\
& 4 & -- & -- & -- & 74.98 & 72.21 \\
& 5 & \checkmark & -- & -- & 75.55 & 72.33 \\
& 5 & \checkmark & \checkmark & -- & 75.65 & 72.28 \\
& 6 & \checkmark & \checkmark & \checkmark & 76.07 & 72.80 \\
\hline

\multirow{4}{*}{\centering Ours} 
& 3 & -- & -- & -- & 71.48 & 69.61 \\
& 3 & -- & \checkmark & -- & 72.35 & 70.71 \\
& 4 & -- & -- & -- & 75.86 & 72.39 \\
& 5 & \checkmark & -- & -- & 76.11 & 73.05 \\
& 5 & \checkmark & \checkmark & -- & 76.34 & 73.22 \\
& 6 & \checkmark & \checkmark & \checkmark & \textbf{77.22} & \textbf{73.45} \\
\hline
\end{tabular}
\end{table}
\FloatBarrier
\subsection{Additional Cross-Dataset Analysis on BRCA}
To further investigate the behavior of contextual guidance beyond the OCELOT benchmark, we additionally evaluated the proposed framework on a BRCA histopathology dataset~\cite{abousamra2021multi1}. Unlike OCELOT, which explicitly models the interaction between tissue-level context and cell-level representations through paired tissue and cellular regions-of-interest, the BRCA setting does not provide explicit tissue-aware contextual supervision. Instead, the experiments focus on the effect of cellness-based localization guidance in a multi-class cell discrimination scenario involving lymphocyte, tumor epithelial, and stromal cells. Table~\ref{tab:brca_analysis} summarizes the experimental results. Existing context-aware approaches already achieve strong performance on this benchmark, demonstrating the effectiveness of contextual modeling for histopathology cell detection. In particular, the recent context-aware detector~\cite{shui2026towards} achieved a Macro F1-score of 0.7201, highlighting the importance of integrating broader contextual information even without explicit tissue-level supervision.

Starting from the RGB-only dual-encoder baseline, we explored different strategies for incorporating cellness guidance. Directly injecting cellness-related information as an additional input signal produced limited improvements and introduced instability due to the noisy nature of teacher-generated localization priors. In contrast, the proposed Cellness Gate consistently improved both localization and classification performance, achieving the best overall Macro F1-score of 0.7258 and the highest Detection F1-score of 0.8562.
\begin{table}[ht]
\centering
\caption{Additional cross-dataset analysis on the BRCA dataset.}
\label{tab:brca_analysis}
\footnotesize
\setlength{\tabcolsep}{4pt}
\begin{tabular}{lccccc}
\toprule
\textbf{Method} & \textbf{Lymph} & \textbf{Tumor} & \textbf{Stromal} & \textbf{Macro F1} & \textbf{Det. F1} \\
\midrule

MCSpatNet~\cite{abousamra2021multi1} 
& 0.6350 
& 0.7850 
& 0.5530 
& 0.6580 
& 0.8490 \\

Context-aware Detector~\cite{shui2026towards}
& 0.7268
& \textbf{0.8349}
& 0.5987
& 0.7201
& -- \\

Ours (RGB + FG )
& 0.7578 
& 0.8014 
& 0.5720 
& 0.7104 
& 0.8317 \\

Ours (RGB) 
& 0.7511 
& 0.8162
& 0.5727 
& 0.7133 
& 0.8279 \\

Ours (Cellness Gate)
& \textbf{0.7631} 
& 0.8149 
& \textbf{0.5992} 
& \textbf{0.7258} 
& \textbf{0.8562} \\

\bottomrule
\end{tabular}
\end{table}
\FloatBarrier
\section{Conclusion}

In this work, we proposed a prior-aware dual-encoder framework for context-aware cell detection in histopathology images. The proposed architecture combines CNN-based local feature modeling with transformer-based global contextual reasoning through a learnable prior-gated fusion mechanism, enabling adaptive integration of contextual priors while reducing the influence of noisy guidance signals.

Experiments on the OCELOT benchmark demonstrated consistent improvements over existing approaches, highlighting the importance of jointly modeling local morphology and global tissue context. Additional experiments on the BRCA dataset further showed that adaptive gated prior integration is more effective than direct prior injection, improving both localization and classification performance even without explicit tissue-level contextual supervision.

Overall, the results suggest that reliable contextual integration plays a critical role in robust computational pathology cell detection. Future work will investigate uncertainty-aware prior modeling and stronger cross-domain generalization across diverse pathological datasets.

\bibliography{z-references}
\bibliographystyle{splncs04}
\end{document}